# Predicting Clinical Events by Combining Static and Dynamic Information using Recurrent Neural Networks

Cristóbal Esteban
Siemens AG and
Ludwig Maximilian
University of Munich
Munich, Germany

Oliver Staeck
Charité
University Hospital of Berlin
Berlin, Germany

Stephan Baier
Siemens AG and
Ludwig Maximilian
University of Munich
Munich, Germany

Yinchong Yang
Siemens AG and
Ludwig Maximilian
University of Munich
Munich, Germany

Volker Tresp
Siemens AG and
Ludwig Maximilian
University of Munich
Munich, Germany

***Abstract*—In clinical data sets we often find static information (e.g. patient gender, blood type, etc.) combined with sequences of data that are recorded during multiple hospital visits (e.g. medications prescribed, tests performed, etc.). Recurrent Neural Networks (RNNs) have proven to be very successful for modelling sequences of data in many areas of Machine Learning. In this work we present an approach based on RNNs, specifically designed for the clinical domain, that combines static and dynamic information in order to predict future events. We work with a database collected in the Charité Hospital in Berlin that contains complete information concerning patients that underwent a kidney transplantation. After the transplantation three main endpoints can occur: rejection of the kidney, loss of the kidney and death of the patient. Our goal is to predict, based on information recorded in the Electronic Health Record of each patient, whether any of those endpoints will occur within the next six or twelve months after each visit to the clinic. We compared different types of RNNs that we developed for this work, with a model based on a Feedforward Neural Network and a Logistic Regression model. We found that the RNN that we developed based on Gated Recurrent Units provides the best performance for this task. We also used the same models for a second task, i.e., next event prediction, and found that here the model based on a Feedforward Neural Network outperformed the other models. Our hypothesis is that long-term dependencies are not as relevant in this task.**

## I. Introduction

As a result of the recent trend towards digitization, a growing amount of information is recorded in clinics and hospitals and therefore the human expert is increasingly overwhelmed. This problem is one reason why Machine Learning (ML) is gaining attention in the medical domain, since ML algorithms can make use of all of the available information to predict the most likely future events that will occur to each individual patient. Physicians can take advantage of these predictions in their decisions which might lead to improved outcomes. ML can also be the basis for a decision support system that provides personalized recommendations for each individual patient (e.g., which is the most suitable medication, which procedure should be applied next, etc.).

It is also worth noticing that the medical data sets are becoming both longer (i.e. we have more samples collected through time) and wider (i.e. we store more variables). Therefore we need to use Machine Learning algorithms capable of capturing complex relationships among a big number of time-evolving variables.

A class of algorithms that can model very complex relationships are Neural Networks, which have proven to be successful in other areas of Machine Learning [1]. Particularly, there is a notable parallelism among the prediction of clinical events and the field of Language Modelling, where Deep Learning, a class of Neural Networks with multiple hidden layers, has also proven to be very successful. One could imagine that each word of a text represents an event. Therefore a text would be a stream of events and the task of Language Modelling would be to predict the next event in the stream. For this reason, we can get inspired by Language Modelling to create models that predict clinical events. However, the medical domain has a set of characteristics that make it an almost unique scenario: multiple events can occur at the same time, there are multiple sequences (i.e. multiple patients), each sequence has an associated set of static variables and both inputs and outputs can be a combination of Boolean variables and real numbers. For these reasons we need to develop approaches specifically designed for the medical use case.

For our work we use a large data set collected from patients that suffered from kidney failure. The data was collected in the Charité hospital in Berlin and it is the largest data collection of its kind in Europe. Once the kidney has failed, patients face a lifelong treatment and periodic visits to the clinic for the rest of their lives. Until the hospital finds a new kidney for the patient, he or she must attend to the clinic multiple times per week in order to receive dialysis, which is a treatment that replaces many of the functions of the kidney. After the transplant has been performed, the patient receives immunosuppressive therapy to avoid the rejection of the transplanted kidney. The patient must be periodically controlled to check the status of the kidney, adjust the treatment and take care of associated diseases, such as those that arise due to the immunosuppressive therapy. This data set started being recorded more than 30 years ago and it is composed of more than 4000 patients that underwent a renal transplantation or are waiting for one. The database has been the basis for many studies in the past [2], [3], [4], [5].

There are a set of endpoints that can occur after a transplantation and it would be very valuable for the physicians to be able to know beforehand when one of these is going to happen. Specifically we will predict whether the patient will die, the transplant will be rejected, or the transplant will be lost. For each visit that a patient makes to the clinic, we will anticipate which of those three events (if any) will occur both within 6 months and 12 months after the visit.

In order to accomplish the prediction of these endpoints, we developed a new model based on Recurrent Neural Networks (RNNs). The main advantage of our model is its ability to combine static and dynamic information from patients. This capability is very important for medical applications, since most of the clinical data sets present some background information about the patients (e.g. gender, blood type, main disease, etc.) combined with dynamic information that is recorded during the multiple visits to the clinic (e.g. results of the laboratory tests, prescribed medications, etc.).

The paper is organized as follows. In the next section we discuss alternative approaches for endpoint predictions. One example is a Feedforward Neural Network that is also able to deal with dynamic and static information. This model will be used as our main benchmark. In Section III we describe details of the nephrology use case and explain why anticipating these endpoints could be of great value for physicians. Section IV introduces the proposed models for this work, starting with a brief overview on RNNs followed by the specific details of our work. In Section V we explain the experimental set ups and present our results. Section VI contains our conclusions and an outlook.

## II. Related Work

Regarding the task of predicting clinical events, Esteban et al. [6] introduced the Temporal Latent Embeddings model which is based on a Feedforward Neural Network. This model outperforms its baselines for the task of predicting which events will be observed next given the previous events recorded (i.e. the goal was to predict which laboratory analyses and medication prescriptions will be observed in the next visit for each patient).

The architecture of this model can be seen in Figure 1. It takes two types of inputs. On the bottom left corner of the picture we can find the input vector that contains the static information of the patients and an aggregation of their medical histories. Therefore, this vector contains the background information of the patient. The rest of the input vectors that we can see in Figure 1 form a set of $n$ vectors that contain the events that occurred in the $n$ steps previous to the one we want to predict.

Each of the input vectors is fed into a representation layer. The representation layer compresses the input vectors into their latent representations, which are vectors composed of real numbers. Then, the $n+1$ latent representations are stacked together horizontally and the resulting vector is used as the input of a Feedforward Neural Network.

Overall, the Temporal Latent Embeddings resembles an $n$-th order Markov model, and due to its architecture, it requires to explicitly define the number of time steps in the past that we want to consider in order to predict the next step. In some scenarios, this constraint can actually be an advantage since many recent papers have shown how attention mechanisms do actually improve the performance of the Neural Networks [8] [9]. The Temporal Latent Embeddings model puts all its attention on the last $n$ samples and therefore it provides an advantage over RNNs for data sets where the events that we want to predict are dependent just on the $n$ previous events. However, in order to capture long term dependencies on the data with this model, we have to aggregate the whole history of each patient in one vector (e.g. computing the mean values of each laboratory measurement), and therefore many long-term dependencies can be lost in this aggregation step (e.g. a very high value in one measurement followed by a very low value).

In recent work, Choi et al. [10] used an RNN with Gated Recurrent Units (GRUs) combined with skip-gram vectors [11] in order to predict diagnosis, medication prescription and procedures. However in this work they follow a standard RNN architecture that takes sequential information as input, whereas the static information of the patients was not integrated into the Neural Network. However, due to its nature, medical data will always contain both static and dynamic features, and therefore it is fundamental to develop algorithms that can combine and exploit both types of data.

Outside of the medical domain, there are multiple examples where RNNs have been successfully applied to sequential data sets in order to predict future events. For example, in Natural Language Processing RNNs are commonly used to predict the next word in a text or even full sentences [12] [13]. However to the best of our knowledge none of them includes both sequential and static information, which is the typical setting in sequences of clinical data.

## III. Kidney Transplantation Endpoints

Chronic kidney disease is a worldwide health burden with increasing prevalence and incidence. It affects multiple organ systems and may result in end-stage renal disease. The growing number of patients with end-stage renal disease requiring dialysis or transplantation is a major challenge for healthcare systems around the world [14]. For most patients, kidney transplantation is the treatment of choice offering lowest morbidity, significant survival benefit, lowest costs and highest quality of life. The main goals after transplantation are the reduction of complications and the increase of long-term graft and patient survival. However, kidney transplant recipients are at high risk of severe complications like rejections, infections, drug toxicity, malignancies and cardiovascular diseases. The majority of patients have to take 5-10 different medications every day during their entire life. Due to complexities of post-transplant management, kidney transplant recipients should remain in life-long specialized care. The medical records of kidney transplant recipients mostly cover a very long history of disease (years to decades) and include a vast number of diagnoses, symptoms, results, medications and laboratory values.

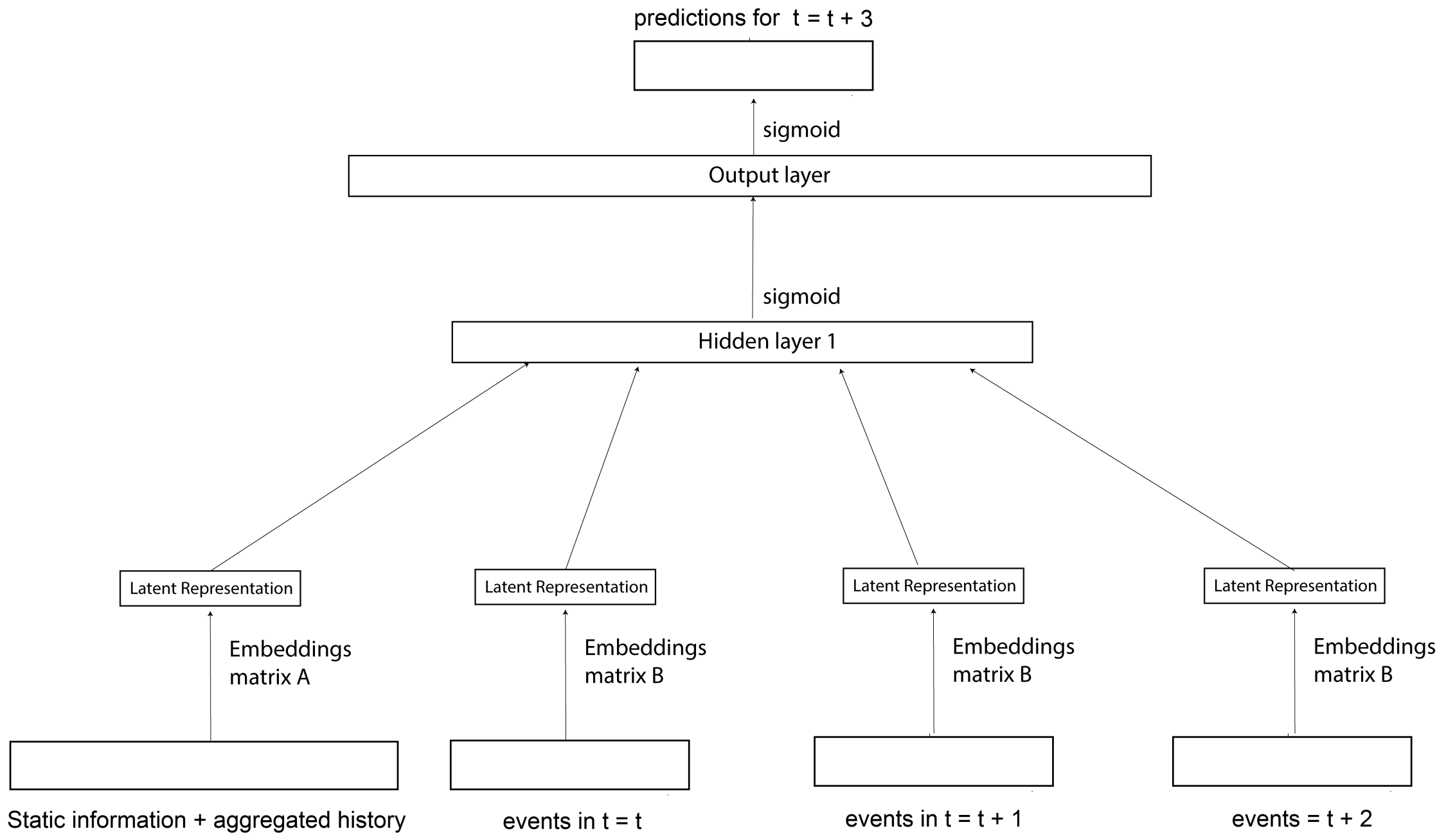

Fig. 1. Temporal Latent Embeddings Model.

Due to this complexity of medical data, decision making is complex and difficult in the clinical practice of kidney transplantation. Considering the load of treating 20-40 patients per day, it is generally not possible for the medical practitioner to review all available medical information during every bed side visit or outpatient consultation. Early prediction of clinical events on the base of current data, as proposed in our proposed solution, can lead to informed decision making on how to best choose future therapy and identify current problems. Thus our proposed solution can help to avoid complications and improve survival of the patient and graft, reduce morbidity and improve health related quality of life.

Assessing the risk of graft failure or death in renal transplant recipients can be crucial for the individual patient management by detecting patients that need intensified medical care. Integrating a computerized tool to predict relevant end points (deteriorating of the graft function, infections, rejections, graft loss, death) on the base of all available medical data may be of great benefit in the clinical routine and provide medical professionals with significant decision support.

Detecting a higher risk of graft failure or death 6-12 months in advance may timely allow identifying toxicities, side effects, interactions of medications, infections, relevant comorbidities and other complications. Early detection permits timely intervention with a chance of improved outcome. Furthermore, predicting future rejections during consultations enables the medical practitioner to change immunosuppressive medications and thus prevent deterioration of the graft function. In summary, computerized prediction of relevant events has the potential to significantly change daily medical practice in patient care.

There are three major endpoints that can happen after a kidney transplantation: rejection of the kidney, graft loss and death of the patient. A rejection means that the body is rejecting the kidney. In such situation, physicians try to fight the rejection with medications and if they are not able to stop the rejection, the kidney will eventually stop working.

Our goal with this work is to predict which of these endpoints (if any) will occur to each patient 6 months and 12 months after each visit to the clinic, given the medical history of the patient. Our predictions are based on information from the patient's medical history, the sequence of medications prescribed for the patient, the sequence of laboratory tests performed together with their results, and static information, as for example age, gender, blood type, weight, primary disease, etc. Therefore each patient is represented as a sequence of events (medications prescribed and laboratory tests performed) combined with static data. We will use both static and dynamic data to predict the explained endpoints.

## IV. Recurrent Neural Networks for Clinical Event Prediction

RNNs are a type of Neural Network where the hidden state of one time step is computed by combining the current input with the hidden state of the previous time step. In this way, RNNs can learn to remember events from the past that are relevant to predict future outcomes. Figure 2 shows the architecture of an RNN.

An advantage of RNNs is that the memory is essentially unlimited, whereas for Feedforward Neural Networks the time window relevant for prediction needs to be specified: RNNs can in principle learn to remember any event (or combination of events) that occur in the life of the patient that is useful to predict future events. This feature can be very valuable when our clinical data set presents such kind of long term dependencies.

Another advantage of using RNNs for learning with sequences is that, given a new patient, we can start predicting future events for such patient right after his or her first visit to the clinic. On the other hand, if we model our data with a Feedforward Neural Network and we have decided to use the $n$ previous visits to predict future events, we will have to wait until we have accumulated at least $n$ visits for a patient to start predicting his or her future events. In some scenarios, this ability of making predictions using a variable amount of previous time steps can be a very useful feature.

More formally, the output of an RNN is computed as:

$$\hat{y}_t = \sigma(W_o h_t) \tag{1}$$

where $W_o$ is a matrix containing the parameters of the model, $h_t$ is a vector containing the hidden state of the Neural Network, $\sigma$ is some differentiable function that is applied element-wise (usually the logistic sigmoid function or the hyperbolic tangent) and $\hat{y}_t$ is a vector containing the predicted output.

Given a sequence of input vectors $x = (x_1, x_2, \cdots, x_T)$, the hidden state of an RNN is computed the following way:

$$h_t = f(h_{t-1}, x_t). \tag{2}$$

We will summarize the most common options for the $f$ function in the next sections.

### A. Standard Recurrent Neural Network

The hidden state in RNNs is updated as:

$$h_t = \sigma(W x_t + U h_{t-1}) \tag{3}$$

where $W$ and $U$ are two matrices that contain the parameters of the model.

In order to compute the value of the hidden state at time $t$, we perform a linear combination of the hidden state of the previous time step $h_{t-1}$ together with the current input to the network $x_t$. In this way the Neural Network can learn to remember specific events observed in the past by encoding them in the hidden state.

As we mentioned earlier, one of our main interests for using RNNs is to capture long-term dependencies. The ability to remember events that occurred a long time ago can be specially useful for some medical applications where events observed at some point in the life of the patient can be very informative about future events that will be observed.

However, it was observed by Bengio et al. [15], that it is not possible for standard RNNs to capture long-term dependencies from very far in the past due to the vanishing gradient problem. The vanishing gradient problem means that, as we propagate the error through the network to earlier time steps, the gradient of such error with respect to the weights of the network will exponentially decay with the depth of the network. This happens because for each additional time step that we go into the past, we multiply the gradient of the error by the derivative of the sigmoid and a matrix composed of very small numbers. The derivative of the sigmoid has a maximum value of 0.25. Therefore, as after a few time steps, we have multiplied the gradient multiple times by very small numbers and therefore it tends to 0. Pascanu et al. recently published a thorough study on the subject [16]. It has been empirically checked that standard RNNs cannot remember events that occurred around 5-10 time steps into the past.

### B. Long Short-Term Memory units

In order to alleviate the gradient vanishing problem, Hochreiter et al. [17] developed a gating mechanism that dictates when and how the hidden state of an RNN has to be updated.

There are different versions with minor modifications regarding the gating mechanism in the long short-term memory units (LSTM) units. We will use in here the ones defined by Graves et al.[18].

$$i_t = \sigma(W_{xi} x_t + W_{hi} h_{t-1} + W_{ci} c_{t-1} + b_i) \tag{4}$$

$$r_t = \sigma(W_{xr} x_t + W_{hr} h_{t-1} + W_{cr} c_{t-1} + b_r) \tag{5}$$

$$c_t = r_t \odot c_{t-1} + i_t \odot \tanh(W_{xc} x_t + W_{hc} h_{t-1} + b_c) \tag{6}$$

$$o_t = \sigma(W_{xo} x_t + W_{ho} h_{t-1} + W_{co} c_{t-1} + b_o) \tag{7}$$

$$h_t = o_t \odot \tanh(c_t) \tag{8}$$

where $\odot$ is the element-wise product and $i$, $r$ and $o$ are the input, forget and output gates, respectively. As it can be seen, the gating mechanism regulates how the current input and the previous hidden state must be combined to generate the new hidden state.

The main differences between this implementation of the LSTM units and the first one published by Hochreiter et al., is that the original implementation did not include forget gates and the activation function was the sigmoid function instead of tanh.

LSTM units have been successfully used many times before [8] [9]. Indeed they have become the de facto standard for RNN implementations. The only downside of LSTM units, when compared with standard RNN updates, is the increase in the number of network parameters, which makes them more computationally expensive. However, their excellent prediction performance makes them the preferred choice in most applications.

### C. Gated Recurrent Units

Another gating mechanism, named Gated Recurrent Units (GRUs), was introduced by Cho et al. [19] with the goal of making each recurrent unit to adaptively capture dependencies of different time scales. We follow the equations as defined by Chung et al. [20]:

$$r_t = \sigma\left(W_r x_t + U_r h_{t-1}\right) \tag{9}$$

$$\tilde{h}_t = \tanh\left(W x_t + U\left(r_t \odot h_{t-1}\right)\right) \tag{10}$$

$$z_t = \sigma\left(W_z x_t + U_z h_{t-1}\right) \tag{11}$$

$$h_t = (1 - z_t) h_{t-1} + z_t \tilde{h}_t \tag{12}$$

where $z$ and $r$ are the update and reset gates respectively.

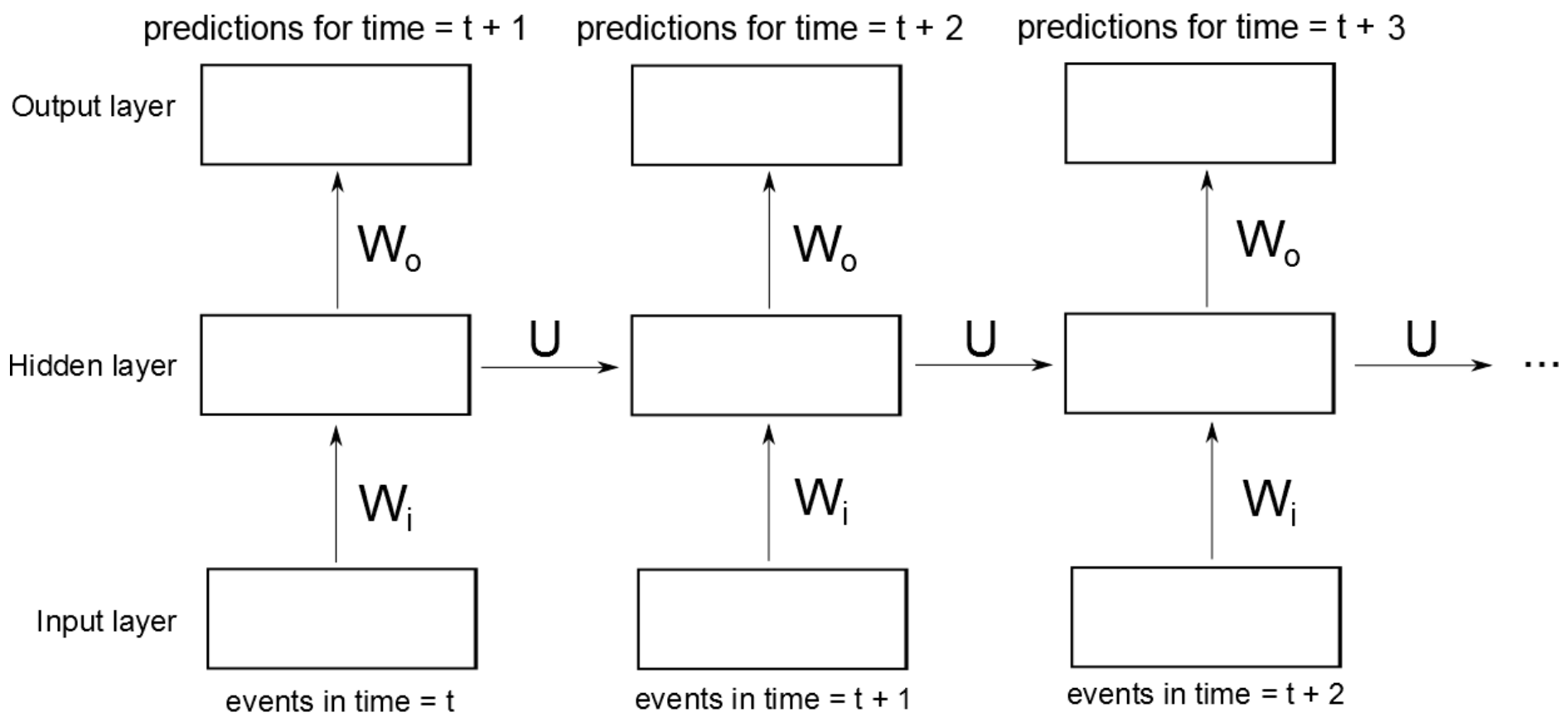

Fig. 2. Recurrent Neural Network.

As it can be seen, GRUs are less complex than LSTM units, and experimentally in many cases perform better.

In [20], Chung et al. compare the performance provided by standard RNNs, LSTM units and GRU units, using multiple data sets. It was shown that GRU units outperform the other models in most cases.

### *D. Combining RNNs with Static Data*

It often happens that Electronic Health Records (EHRs) contain both dynamic information (i.e. new data is recorded every time a patient visits the clinic or hospital) and static or slowly changing information (e.g. gender, blood type, age, etc.).

Therefore, we modified the RNN architecture to include static information of the patients, such as gender, blood type, cause of the kidney failure, etc. The modified architecture is depicted in Figure 3.

As it can be seen, we process the static information on an independent Feedforward Neural Network whereas we process the dynamic information with an RNN. Afterwards we concatenate the hidden states of both networks and provide this information to the output layer.

We feed our network with the latent representation of the inputs, as it is often done in Natural Language Processing [21] [13]. In this case we compute such latent representation applying a linear transformation to the raw input.

More formally, we first compute the latent representation of the input data as

$$\tilde{x}_i^e = A \tilde{x}_i \tag{13}$$

$$x_{t,i}^e = B x_{t,i} \tag{14}$$

where $\tilde{x}_i$ is a vector containing the static information for patient $i$ being $\tilde{x}_i^e$ its latent representation, and $x_{t,i}$ is a vector containing the information recorded during the visit made by patient $i$ at time $t$, where $x_{t,i}^e$ is its latent representation.

Then we compute the hidden state of the static part of the network as

$$\tilde{h}_i = \tilde{f}(\tilde{x}_i^e) \tag{15}$$

where we are currently using the input and hidden layers of a Feedforward Neural Network as $\tilde{f}$.

In order to compute the hidden state of the recurrent part of the network we update:

$$h_{t,i} = f(x_{t,i}^e, h_{t-1,i}) \tag{16}$$

where $f$ can be any of the update functions explained above (standard RNN, LSTM or GRU).

Finally, we use both hidden states in order to predict our target:

$$\hat{y}_{t,i} = g(\tilde{h}_i, h_{t,i}). \tag{17}$$

In this work we concatenate $\tilde{h}_i$ and $h_{t,i}$, and then the function $g$ is computed as

$$g = \sigma(W_o(\tilde{h}_i, h_{t,i}) + b). \tag{18}$$

We derive a cost function based on the Bernoulli likelihood function, also known as Binary Cross Entropy, which has the form

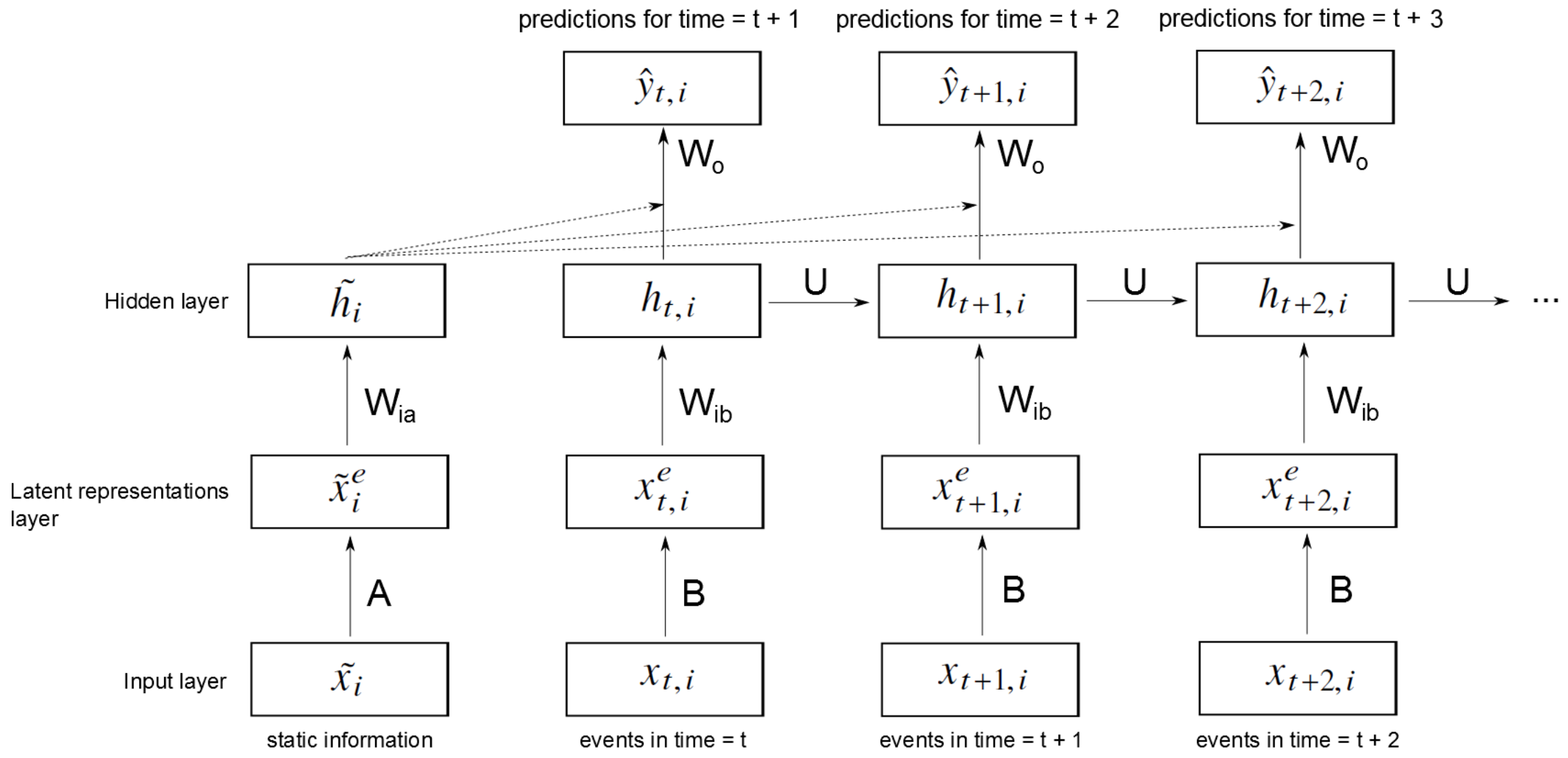

Fig. 3. Recurrent Neural Network with static information. $i$ stands for the index of the patient.

$$\text{cost}(A, B, W, U) = \sum_{t,i \in \text{Tr}} -y_{t,i} \log(\hat{y}_{t,i}) - (1 - y_{t,i}) \log(1 - \hat{y}_{t,i}) \quad (19)$$

where Tr stands for the training data set, $y_{t,i}$ stands for the true target data and $\hat{y}_{t,i}$ stands for the predicted data.

It is worth mentioning that we tried other architectures for combining static data with dynamic data using RNNs. For example we experimented with the approach of using an RNN to compute a latent representation of the whole history of the patient, in order to use such representation together with the latent representations of the last $n$ visits as the inputs of a Feedforward Neural Network. We thought it could be an interesting approach since the network would put a higher attention on the most recent visits and would still have access to the full history of the patient. We also tried an attention mechanism similar to the ones shown in [8] [9], which also has the advantage of providing an interpretable model (i.e. it provides information about why a prediction was made). However none of these approaches was able to improve the performance of the models presented in this article.

## V. Experiments

### A. Data Pre-processing and Experimental Setup

We have three groups of variables in our data set: endpoints, prescribed medications and laboratory results. Both endpoints and prescribed medications are binary data. On the other hand, the laboratory results are a set of real numbers.

Not every possible laboratory measurement is performed each time a patient visits the clinic (e.g. some times a doctor may order to measure calcium if it is suspected that the patient might have low calcium, otherwise the doctor will not order such measurement). In order to deal with such missing data, we experimented with mean imputation and median imputation. Also, we experimented with both scaling and normalizing the data before inputting it to the Neural Network.

It turned out that encoding the laboratory measurements in a binary way by representing each of them as three event types as in [6], i.e., LabValueHigh, LabValueNormal and LabValueLow, provided a better predictive performance than the approach of doing mean or median imputation and normalizing or scaling the data. This improvement was around 5% for the area under the ROC curve score and 3% for the area under the precision-recall curve score. In order to encode the laboratory measurements into High, Normal and Low values, we calculated the mean and standard deviation for each of them. Then, measurements greater than the mean plus the standard deviation were encoded as high, values below the mean minus the standard deviation were encoded as low, and values in between were encoded as normal. If a certain measurement was not done, its corresponding three events are all set to 0, which removes the need of imputing missing data. We believe that this improved performance provided by the discretization of the inputs, is due to the increase in the number parameters of the model and due to the fact that with this strategy we remove the need of doing data imputation, which can add a significant noise to the data set.

The final aspect of the pre-processed data that we will use as input for our models can be found in Figure 4, where each row represents a visit to the clinic. The target data (i.e. the data we want to predict) will be a matrix that contains a row composed of 6 binary variables for each row on the input matrix. This 6 binary variables specify which of the 3 endpoints (if any) occurred 6 and 12 months after each visit. The possible endpoints are: kidney rejection, kidney loss and death of the patient.

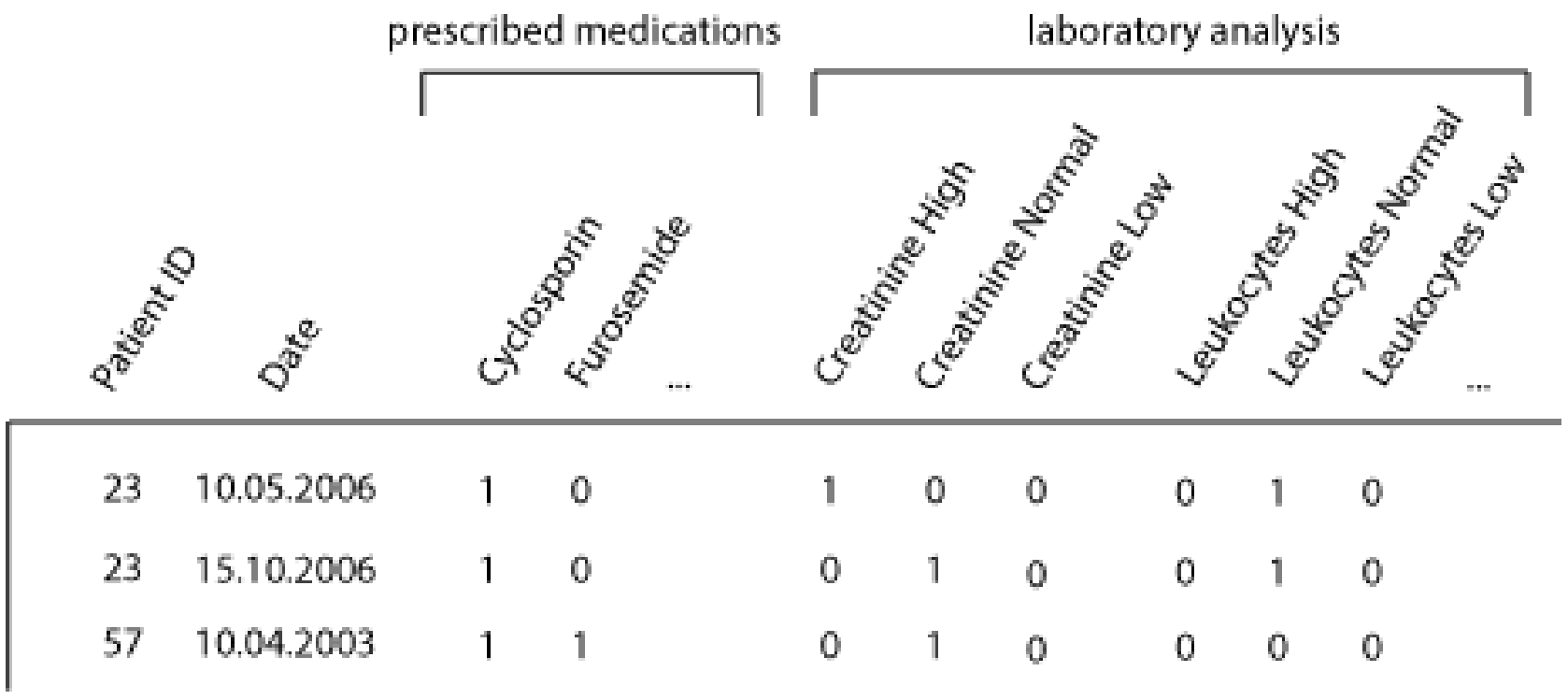

Fig. 4. Sample of pre-processed data that we use as input for our models. Each row represents a visit to the clinic.

Another option for performing this binary encoding would be to somehow normalize the measured values using demographic data of each patient (e.g. gender, age, weight) or to use normal and limit values according to the medical literature. We will explore those options in future work.

In this article we consider events from the data set that occurred on the year 2005 and onwards due to the improvement of the data quality from that year on. After cleaning the data the total number of patients is 2061, which in total have made 193111 visits to the clinic. The density of end-points (target matrix) is 7.3%, and 38.4% of the patients have suffered at least one end-point event.

The dynamic information that is generated on each visit to the clinic is composed of 1061 medications that can be prescribed and 1835 substances that can be measured in the laboratory. Thus, given that we encode each laboratory measurement into three binary variables as explained earlier, the dynamic information is composed of a total of 6566 variables. On the other hand, the static information is composed of 342 features that remain constant for each patient.

The model contains several hyperparameters that need to be optimized. The most relevant ones are the rank of the latent representations, the number of hidden units in the Neural Network, the learning rate and the dropout regularization parameter [7]. Besides we will test our models with two optimization algorithms, which are Adagrad [22] and RMSProp [23].

In order to fit these hyperparameters, we randomly split the data into three subsets: 60% of the patients were assigned to the training set, 20% of the patients were assigned to the validation set and another 20% to the test set. Under this configuration, we evaluate the performance of the model by predicting the future events of patients that the model has never seen before, and therefore increasing the difficulty of the task.

When comparing the performance of these models, we report for each model the mean area under the Precision-Recall curve (AUPRC) and mean area under Receiver Operating Characteristics curve (AUROC) together with their associated standard errors after repeating each experiment five times with different random splits of the data. We made sure that these five random splits were identical for each model.

It is worth noting that in use cases where the target event we want to predict is very infrequent, and where we are more interested in knowing when such event is going to happen (as opposed to when it is not going to happen), then the AUPRC is a more interesting score to evaluate the quality of the predictions than the AUROC. This is because getting high sensitivity and specificity can be fairly easy in these problems. However, obtaining a high precision in the predictions is a very hard challenge. Indeed, we will show in the next section how making random predictions provides a very low AUPRC, much lower than the AUROC for random predictions. Therefore, the most interesting score to compare the models presented in this work is the AUPRC.

We will also report results on the Logistic Regression, since it provided the second best performance in the task of predicting sequences of clinical data in [6].

### B. Results

Table I shows the results of predicting endpoints without considering different types of them (i.e. we concatenate all the predictions of the set and evaluate all of them together). "GRU + static", "LSTM + static" and "RNN + static" stand for the architectures presented in this paper that combine an RNN with static information. TLE stands for the Temporal Latent Embeddings model [6]. Random stands for the scores obtained when doing random predictions.

We can see how the recurrent models outperform the other models both in the AUROC score and AUPRC score. The best performance is achieved by the GRUs with an AUPRC of 0.345 and an AUROC of 0.833. The AUPRC scores are pretty low compared to its maximum possible value (i.e. 1), but are fairly good compared to the random baseline of 0.073, which is that low due to the high sparsity of the data.

Since we repeated the experiment five times with different splits of the data, we have slight variations on the configuration of the winning model. However, the most repeated configuration was composed of 100 hidden units, a rank size of 50 for the latent representation, a dropout rate of 0.1, a learning rate of 0.1 and the Adagrad optimization algorithm.

TABLE I
SCORES FOR ENDPOINT PREDICTION. AUPRC STANDS FOR AREA UNDER PRECISION-RECALL CURVE. AUROC STANDS FOR AREA UNDER ROC CURVE.

| | AUPRC | AUROC |
|---|---|---|
| GRU + static | 0.345 ± 0.013 | 0.833 ± 0.006 |
| LSTM + static | 0.330 ± 0.014 | 0.826 ± 0.006 |
| RNN + static | 0.319 ± 0.012 | 0.822 ± 0.006 |
| TLE | 0.313 ± 0.010 | 0.821 ± 0.005 |
| Logistic Regression | 0.299 ± 0.009 | 0.808 ± 0.005 |
| Random | 0.073 ± 0.002 | 0.5 |

In Table II we show the performance achieved for each specific endpoint. It can be seen how the it gets the best score for predicting the death of a patient whereas it obtains the worse AUPRC in the task of predicting kidney loss within the next 6 months.

TABLE II
GRU + STATIC SCORES FOR ENDPOINT PREDICTION. AUPRC STANDS FOR AREA UNDER PRECISION-RECALL CURVE. AUROC STANDS FOR AREA UNDER ROC CURVE.

| | AUPRC | AUROC |
|---|---|---|
| Rejection 6 months | 0.234 ± 0.010 | 0.778 ± 0.006 |
| Rejection 12 months | 0.279 ± 0.014 | 0.768 ± 0.009 |
| Loss 6 months | 0.167 ± 0.017 | 0.821 ± 0.009 |
| Loss 12 months | 0.223 ± 0.019 | 0.814 ± 0.009 |
| Death 6 months | 0.467 ± 0.018 | 0.890 ± 0.004 |
| Death 12 months | 0.465 ± 0.020 | 0.861 ± 0.004 |

### C. Additional experiments

As we mentioned earlier, the Temporal Latent Embeddings model outperformed the baselines presented in [6] for the task of predicting the events that will be observed for each patient in his or her next visit to the clinic (i.e. which laboratory analyses will be made next, which results will be obtained in such analyses and what medications will be prescribed next). However none of those baselines were based on RNNs.

Thus we reproduced the experiments presented in [6] including the models based on RNNs introduced in this article. Table III shows the result of such experiment, where we can appreciate that the Temporal Latent Embeddings model still provides better scores than the other models. We also included in Table III an entry named “Static embeddings” which corresponds to the predictions made with a Feedforward Neural Network using just the static information of each patient. We hypothesize that the reason that Temporal Latent Embeddings model provides the best performance is due to the lack of complex long term dependencies that are relevant for this task. Thus it would be an advantage to use a model that puts all the attention on the most recent events in situations where all the relevant information to predict the target was recorded during the previous *n* visits.

TABLE III
SCORES FOR NEXT VISIT PREDICTION. AUPRC STANDS FOR AREA UNDER PRECISION-RECALL CURVE. AUROC STANDS FOR AREA UNDER ROC CURVE.

| | AUPRC | AUROC | |
|---|---|---|---|
| TLE | 0.584 ± 0.0011 | 0.978 ± 0.0001 | |
| LSTM + static | 0.571 ± 0.0048 | 0.975 ± 0.0002 | |
| GRU + static | 0.566 ± 0.0034 | 0.975 ± 0.0002 | |
| Static embeddings | 0.487 ± 0.0016 | 0.974 ± 0.0002 | |
| Random | 0.011 ± 0.0001 | 0.5 | - |

## VI. CONCLUSION

We developed and compared novel algorithms based on RNNs that are capable of combining both static and dynamic clinical variables, in order to solve the task of predicting endpoints on patients that underwent a kidney transplantation. This is an application that will provide critical information to physicians and will support them to make better decisions.

We found that an RNN with GRUs combined with a Feedforward Neural Network provides the best score for this task.

We also compared these recurrent models with other models for the task of predicting future medications and laboratory results. We found that for such use case the RNNs do not outperform a model based on a Feedforward Neural Network. We hypothesize that this is due to the lack of complex long term dependencies in the data that are relevant for this task, and therefore it is an advantage to use a model that puts all the attention to the most recent events.

We also found that binary encoding input variables that are composed of real numbers provides a better performance than normalizing the input data and performing imputation to deal with missing data.

## VII. FUTURE WORK

We explained the gradient vanishing problem in Section IV, and later showed how LSTM and GRU units solve this problem. Interestingly, Deep Feedforward Neural Networks also suffer from this problem, also due to the several multiplications by the derivative of the sigmoid (or hyperbolic tangent) function that are accumulated in the first layers of the network. Recently there have been some efforts to solve this problem with a gating mechanism [24], similar to what is done in RNNs. However, the most common solution to solve this problem in Feedforward Neural Networks is to use Rectified Linear Units [25], whose derivative is 1 for positive input values, and therefore mitigate the gradient vanishing problem. It turns out that there have been also efforts to train standard RNNs with Rectified Linear Units in order to prevent the gradient vanishing problem, without the extra computational cost of the gating mechanisms. The results of this approach seems promising and we will add this option to our benchmark, as part of future work.

The key feature of a Clinical Decision Support System is its ability to consider as much patient information as possible and combine it in a meaningful and scalable way to predict future events. We are already capable of combining static information with a sequence of structured data. We plan to integrate more sources of data into our model with the goal of improving the quality of our predictions.

Also, having more powerful models that can deal with more data and more complex data, will allow us to tackle more difficult problems. For example, in the second experiment we predict prescribed medications, but we do not provide information regarding the doses of medications or the intake patterns. As our models improve, we will try to predict this kind of complex targets.

Finally, concerning our specific data set composed of patients that suffered from kidney failure, there is a lot of information that we are not using yet, as for example biopsies and its results. We will keep adding to our models additional sources of information.